%% file: root.tex
\documentclass[letterpaper, 10 pt, conference]{ieeeconf}  

\let\labelindent\relax
\usepackage[utf8]{inputenc} 
\usepackage[T1]{fontenc}    
\usepackage{hyperref}       
\usepackage{url}            
\usepackage{booktabs}       
\usepackage{amsfonts}       
\usepackage{nicefrac}       
\usepackage{microtype}      
\usepackage{amsmath}
\usepackage{enumitem}
\usepackage{subcaption}
\include{macros}

\IEEEoverridecommandlockouts                              

\title{\LARGE \bf Probabilistic Safety Programs}
\author{Ashish Kapoor, Debadeepta Dey and Shital Shah\\
        Microsoft Research, One Microsoft Way, Redmond WA 98052 USA\\
        \tt\{akapoor, dedey, shitals\}@microsoft.com}

\begin{document}
\maketitle
\thispagestyle{empty}
\pagestyle{empty}

\begin{abstract}
Achieving safe control under uncertainty is a key problem that needs to be tackled for enabling real-world autonomous robots and cyber-physical systems. This paper
introduces Probabilistic Safety Programs (PSP) that embed both the uncertainty in the environment as well as invariants that determine safety parameters.
The goal of these PSPs is to evaluate future actions or trajectories and determine how likely it is that the system will stay safe under uncertainty. We propose to perform these evaluations by first compiling the PSP to a graphical model then using a fast variational inference algorithm. We highlight the efficacy of the framework on the task of safe control of quadrotors and autonomous vehicles in dynamic environments.
\end{abstract}

\section{Introduction}
Real-world deployment of autonomous robots and cyber-physical systems (CPS) requires not only reasoning about the uncertain environment, but also using those inferences about the state of the world for {\em safely} completing the assigned mission. For example, autonomous vehicles can be instrumented with various sensors, such as lidar, radar, cameras etc \cite{urmson2008autonomous}. The safety reasoning task then is to consider all the observed data and then determine if the future sequence of actions will yield safe progress towards the goal. However, reasoning about safety of such future actions or control inputs is non-trivial as it requires integrating over all plausible configurations of the environment that would explain the observed data.

We introduce Probabilistic Safety Programs (PSP), that allow inferences about safety of a cyber-physical system in an uncertain environment given a future course of actions. Specifically, a PSP is a stochastic program \cite{GMR+08,GHNR14,MSS09,kulkarni2015picture,BLOG} that allows us to both 1) consider uncertainty in the environment and 2) conditioned on the environment express safety invariants that need to hold. Intuitively, a single run of the PSP draws a sample from the distribution of plausible worlds, and then evaluates if the safety conditions holds for the proposed action sequence. Such probabilistic programs have the advantage of embedding a complex set of graphical models that describe the environment. Furthermore, the rich set of programming constructs allow a wide variety of expressions that describe the safety invariants.

The utility of PSPs can be demonstrated by combining them with existing planners and or control policies. For example, receding horizon planners \cite{howard2010receding} can use PSPs to check for safety of the proposed trajectories. Similarly, Rapidly Exploring Random Trees (RRT/RRT*) \cite{lavalle2000rapidly,karaman2011sampling} style planners can use PSPs to prune those paths that do not meet the safety requirements. All such real-time applications need an efficient inference algorithm. Most of the existing work on Probabilistic Programming relies on sampling based methodologies \cite{GHNR14,WGSS11,SWSG15,WVM14}; consequently, many of the existing inference algorithms are impractical for real-time applications on robots and CPS.

This paper alleviates these computational issues via a very fast variational inference algorithm for PSPs. The key aspect enabling such fast inference is the fact that there exists an equivalent graphical model representation of the underlying semantic structure of PSPs. We also provide an algorithm to recover such equivalent representation. Consequently instead of resorting to sampling, we use fast approximate Bayesian inference.

We demonstrate Probabilistic Safety Programs on three different scenarios: obstacle avoidance with quadrotors, safe battery management and safe maneuvering of autonomous vehicles in dynamic environments. We also show how Probabilistic Safety Programs can help achieve safe performance efficiently when embedded inside an off-the-shelf planning algorithm.

\section{Probabilistic Safety Programs}
Our work on Probabilistic Safety Programs builds upon the prior research in Probabilistic Signal Temporal Logic (PrSTL) \cite{sadighPrSTL2015}, and also incorporates ideas from approximate Bayesian inference for fast computation necessary for real-time implementation. The PrSTL framework was designed for safe controller synthesis in a hybrid dynamic system, where the safety invariants are defined via a distribution of logical expressions that operate on real-valued, dense-time signals. These signals could be functions of the robot state, environment and other safety parameters. The safe controller synthesis then is reduced to a constraint optimization problem, where the mixed-integer constraints are generated from PrSTL specifications. However, optimization under such mixed-integer constraints is computationally intensive and thus real-time implementation of such strategies is non-trivial. Furthermore, the prior work on PrSTL does not easily extend to more complex planning problems, especially in cases where the overall objective function cannot be explicitly expressed. This paper alleviates these problems by considering the probabilistic programming paradigm that enable rich sets of safety expressions under stochasticity and allow for fast inference for real-time applications. Furthermore, such efficient inferential framework also enables embedding of these probabilistic programs as fast safety checks in a wide variety of existing planners and policy generators.

\begin{table}
\small{
\begin{tabular}[t]{llll}
x & $\in$ & Vars &\\
\bf{uop} & ::= & ... & C unary operators\\
\bf{bop} & ::= & ... & C binary operators\\
\bf{$\phi$, $\psi$} & ::= & ... & logical formula\\
&&&\\
$\mathcal{E}$ & ::= & & expression\\
& & | $x$ & variable\\
& & | $c$ & constant\\
& & | ${\mathcal E}_1$ \bf{bop} ${\mathcal E}_2$ & binary operation\\
& & | \bf{uop} ${\mathcal E}$ & unary operation\\
&&&\\
$\mathcal{S}$ & ::= & & statement\\
& & | $x=\mathcal{E}$ & deterministic assignment\\
& & | $x\sim\text{Dist}(\bar{\theta})$ & probabilistic assignment\\
& & | ${\mathcal S}_1$; ${\mathcal S}_2$ & sequential composition\\
& & | For $x = c_1 \mbox{ to } c_2$ & For loop\\
& & | do $\mathcal{S}$ & \\
&&&\\
${\mathcal P}$ & ::= & $\mathcal{S}$ returns $(\mathcal{E})$ & program \\
\end{tabular}}
\caption{Syntax of Probabilistic Safety Programs. The syntax representation is borrowed from \cite{GHNR14}.}
\end{table}

We chose to define Probabilistic Safety Programs using a restricted version of $\text{PROB}$ \cite{GHNR14}, which is a C-like imperative probabilistic programming language. The syntax to define the PSPs is shown in Table 1, and it includes the definition of expressions either as taking deterministic value or a stochastic one.
There are three key differences in the syntax of PSPs with that of $\text{PROB}$.
\begin{enumerate}
\item{Replace \texttt{while} with \texttt{for} loop where the iteration count can be determined at the compilation.}
\item{The syntax does not allow \texttt{if then else} statements.}
\item{Only exponential families are allowed in the probabilistic assignments.}
\end{enumerate}

The key reason behind these modifications is the need for computational efficiency. Specifically, replacing the \texttt{while} loop with a \texttt{for} loop with pre-specified iteration allows for a static unrolling of the program at the time of compilation. Similarly absence of any \texttt{if then} statements means that the sequence of evaluations can be pre-determined. The big consequence of this is that we can represent the probabilistic program as a graphical model, over which we then can run inference algorithms. Finally, only allowing for exponential families in the probabilistic assignments means that fast approximate inference methods such as variational methods can be used easily.

We would like to point out that even with this restricted syntax, PSPs allow for fairly expressive safety specifications. In fact, it is easy to show that  PSPs subsume PrSTL. The proof follows from the observation that PSPs allow for all the logical operators used in PrSTL. Furthermore, the temporal constructs of Globally, Eventually and Until defined in PrSTL can be trivially implemented in a \texttt{for} loop.

Given the capabilities that PSP provides, we can formally define the temporal properties over uncertainties that are present in sensors and classifiers of the system. For example, we can write Probabilistic Safety Program that checks if the output of a Bayesian linear predictor would lie in a desired range for time steps in the future. This is illustrated in the example below:

\noindent{\bf Example 1:} Consider the problem of obstacle avoidance for a robot. Assume that the robot has access to a Bayesian linear predictor ${\bf w} \sim \mathcal{N}({\mathbf \mu}, {\mathbf \Sigma})$ that determines if an obstacle exists at a location ${\bf x}$ via sign$({\bf w}^T {\bf x})$. The corresponding probabilistic safety program\footnote{While the arrays are not explicitly defined in the PSP syntax, we allow slight abuse of notation for the purpose of expositional clarity. The same program can easily be written without arrays.} is depicted in Figure 1. Specifically, inputs to the program are $2$D vector \texttt{x} of $10$ future locations, the mean \texttt{Mu} and the variance \texttt{Sigma} of the Bayesian classifier. The program first draws a sample from the Gaussian distribution of the linear predictors and then for all future locations checks if the safety condition holds. The program returns true only if the dot product of the sampled classifier would predict no obstacle condition for every location.
\begin{figure}[t]
{\scriptsize
\begin{verbatim}
bool AvoidObstacle(double[10, 2] x, double[2] Mu,
                               double[2,2] Sigma)
{
  //Sample the Bayesian linear obstacle classifier
  w = Gaussian(Mu, Sigma);

  bool isSafe = True;
  for (int i = 0; i < x.GetLength(0); i++)
   {
      //Safety invariant for obstacle avoidance
      bool ClearOfObstacles =
          ((w[0]*x[i,0] + w[1]*x[i,1]) > 0);
      isSafe = isSafe && ClearOfObstacles;
   }

 return isSafe;
}
\end{verbatim}}
\caption{Probabilistic Safety Program for Example 1 - Obstacle Avoidance.}
\end{figure}

Given this probabilistic program, we are now interested in the query of the form: what's the probability that the proposed trajectory defined by the set of future locations is safe. Formally, we would like to evaluate $Pr(\texttt{AvoidObstacle(x, Mu, Sigma)} = \mbox{True})$ and if this quantity is greater than a threshold then we deem the trajectory safe. Efficient computation of these quantities is one of the big technical challenges which we show how to tackle next.

Finally, we would like to point out that as the system begins to traverse the trajectory, it has the opportunity to observe new data and update its beliefs over the parameters of the probability distribution. Such data observations and belief updates can naturally be written as probabilistic program constructs within the same framework.

\begin{figure}
{\scriptsize
\begin{verbatim}
bool BatteryAwareFlight(double[] height,
    double[] logbatteryLevel, double variance,
    double heightThresh, double batteryThresh)
{
  bool isSafe = True;
  for (int i = 0; i < height.GetLength(0) - 3; i++)
  {
      //Check if near future steps has high alt. flight
      bool flyHigh = False;
      for (int j = i; j < i + 3; j++)
           flyHigh = flyHigh || (height[j] > heightThresh);

      //Sample battrylevel from the provided distribution
      double batteryNow =
             Gaussian(logbatteryLevel[i], i*variance);
      bool batteryGood = (batteryNow > batteryThresh);

      //Safe invariant that require that high alt. flight
      // requires a healthy battery level
      isSafe = isSafe && (!flyHigh || batteryGood);
  }

 return isSafe;
}
\end{verbatim}}
\caption{Probabilistic Safety Program for Example 2 - Battery Aware Quadrotor Flight.}
\end{figure}

\section{Efficient Inference}
The task of inference in Probabilistic Programs is very challenging. The traditional sampling methods including rejection sampling, likelihood weighing algorithm, and Markov Chain Monte Carlo (MCMC) algorithms could be readily applied to PSPs. However, we note that there are a couple of characteristics in case of Probabilistic Safety Programs that can help with this challenge. First, we are only interested in inferring the probability that a PSP would return true. This is in stark contrast with traditional probabilistic programs where general purpose probabilistic queries can be made. Second, as described in the previous section, the syntax of PSPs allow fully unrolling the code at the compilation time. Consequently it is possible to represent each PSP as a graphical model. Given this graphical model, then existing approximate inference algorithms can be readily applied to infer the required query. We describe these two steps below in detail:

\noindent{\bf Recovering the Graphical Model:} The key idea here is to exploit static analysis to first unroll the program and then induce a graphical model based on the atomic expressions. Specifically, we instantiate a variable in the macro-expanded program for every conceived assignment and binary and unary operations in the original program. The expanded program now is simply a sequence that consists only of either the assignments or unary and binary operations on the instantiated variables.

Once the unrolling is done we then induce a graphical model by spawning a node $\theta$ for every line of code in the expanded program that consists of operation over any of the random variables. Note that $\theta$ directly corresponds to the variable that occurs on the lefthand side of the expression. We then add a directed edge from a node $\theta_i$ to $\theta_j$ if the variable corresponding to $\theta_i$ appears in the righthand side of the line corresponding to $\theta_j$.

Once the edges have been instantiated, we then assign a conditional probability table for each of the nodes. In order to do this, we consider all the incoming edges to a node $\theta$. Note that due to the structure of the expansion, any node can have at most two incoming edges. If the node $\theta$ and all its parents are boolean, then the conditional probability table for that node is completely determined by the corresponding logical formula. If either $\theta$ or any of its parents are continuous then we simply keep track of the dependency and address it during the approximate inference phase.

\begin{figure}
{\scriptsize
\begin{verbatim}
bool AvoidCarCrash(double[] x, double[] y, double[] t,
  double mu_x, double mu_y, double mu_sx, double mu_sy,
  double sigma_sq, double Thresh)
{
  //Sample location and velocities for the other vehicle
  x_other  = Gaussian(mu_x, sigma_sq);
  y_other  = Gaussian(mu_y, sigma_sq);
  sx_other = Gaussian(mu_sx, sigma_sq);
  sy_other = Gaussian(mu_sy, sigma_sq);

  bool isSafe = True;
  for (int i = 0; i < x.GetLength(0); i++)
  {
    //Compute distances to the ego vehicle at each step
    Xdistance = x[i] - (x_other + time[i]*sx_other);
    Ydistance = y[i] - (y_other + time[i]*sy_other);

    //Safety invariants that require min thresh distance
    SafeInX = (Xdistance > Thresh) || (Xdistance < -Thresh);
    SafeInY = (Ydistance > Thresh) || (Ydistance < -Thresh);
    isSafeNow = (SafeInX || SafeInY)

    isSafe = isSafe && isSafeNow;
  }

 return isSafe;
}
\end{verbatim}}
\caption{Probabilistic Safety Program for Example 3 - Collision Avoidance in Autonomous Vehicles.}
\end{figure}

\noindent{\bf Approximate Inference:} The recovered graphical model has both boolean as well as continuous nodes, thus, inference is non-trivial in this case. Note that the program always evaluates to a boolean variable, consequently it is fairly straight forward to show that all the continuous parents have a boolean node as their child. We utilize this observation at runtime, where we marginalize all the continuous nodes upto their discrete children. Once this marginalization happens, the resulting graph is a fully-directed graph with only boolean nodes and their well-defined conditional probability tables.
Formally, each the marginalization operation upto a discrete node $\theta$ can be written as:
\begin{equation}
Pr(\theta = \mbox{True}) \sim \int_{\phi \in \mbox{Asc}(\theta) \cap \mathcal{P}}{f(\phi)}.
\end{equation}
Here $\mbox{Asc}(\theta)$ denotes the set of all the ascendant nodes and the set $\mathcal{P}$ denotes those nodes that correspond to a direct probabilistic assignment in the code. Since $\phi$ arises due to the exponential family in our framework we compute the quantity in equation (1) via approximate variational inference. Once all such quantities are computed, the rest of the remaining graph is a simply directed acyclic graph with boolean variable, thus running inference on the remaining graph is straightforward.

Since the variational approximation lower bounds the integral the resulting approximation has an appealing property of maintaining safety when the sequence of logical relations to the final output does not contain any negation of $\theta$. Intuitively, since the approximation $\hat{Pr}(\theta = \mbox{True}) <= Pr(\theta = \mbox{True})$, the trajectories that are deemed safe with approximation will be safe when there is no negation of $\theta$ when computing the final output of the program.


\begin{figure*}[t]
\centering
\begin{subfigure}[b]{0.3\textwidth}
	\includegraphics[width=\textwidth]{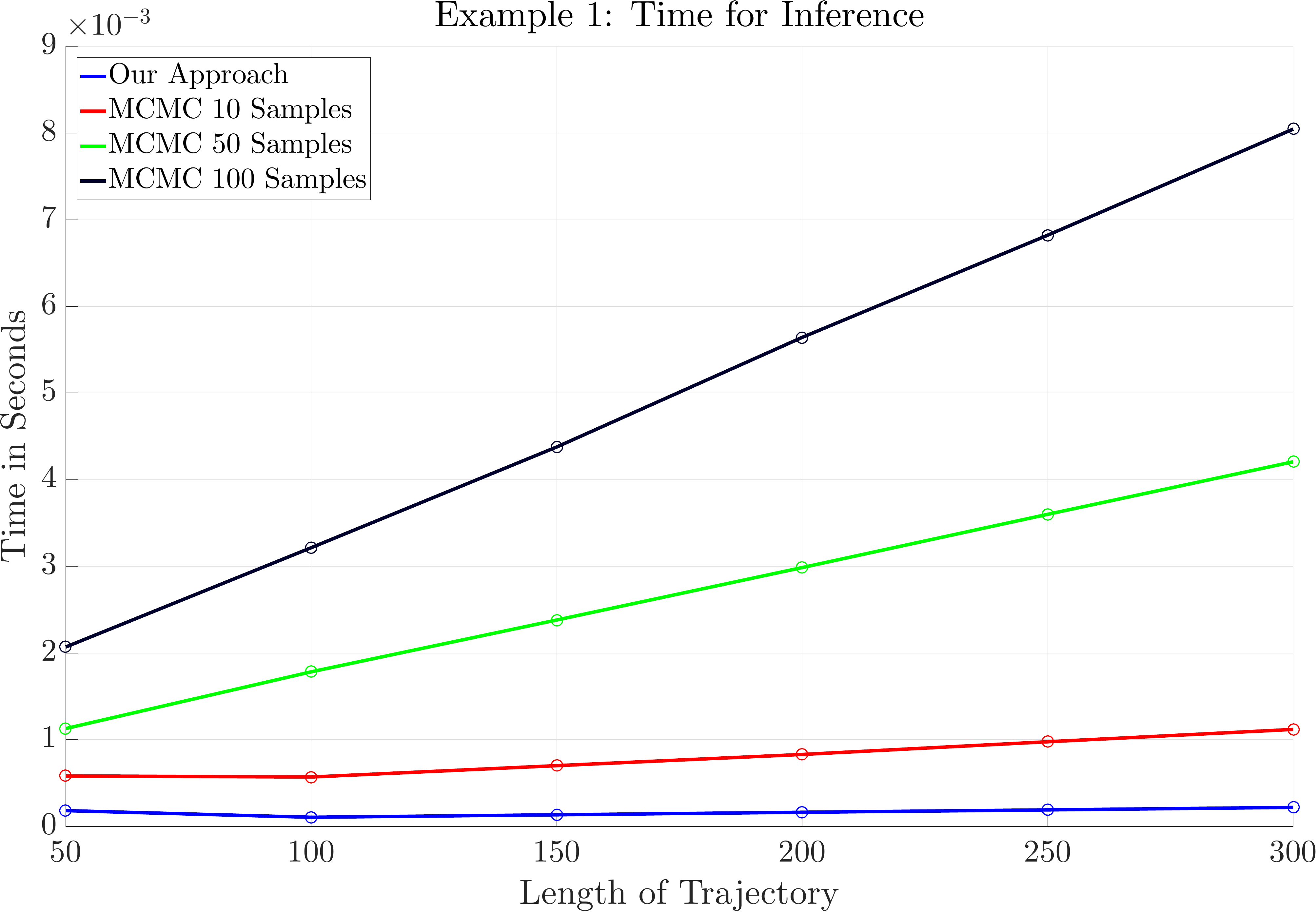}
	\label{blah1}
\end{subfigure}
~
\begin{subfigure}[b]{0.3\textwidth}
	\includegraphics[width=\textwidth]{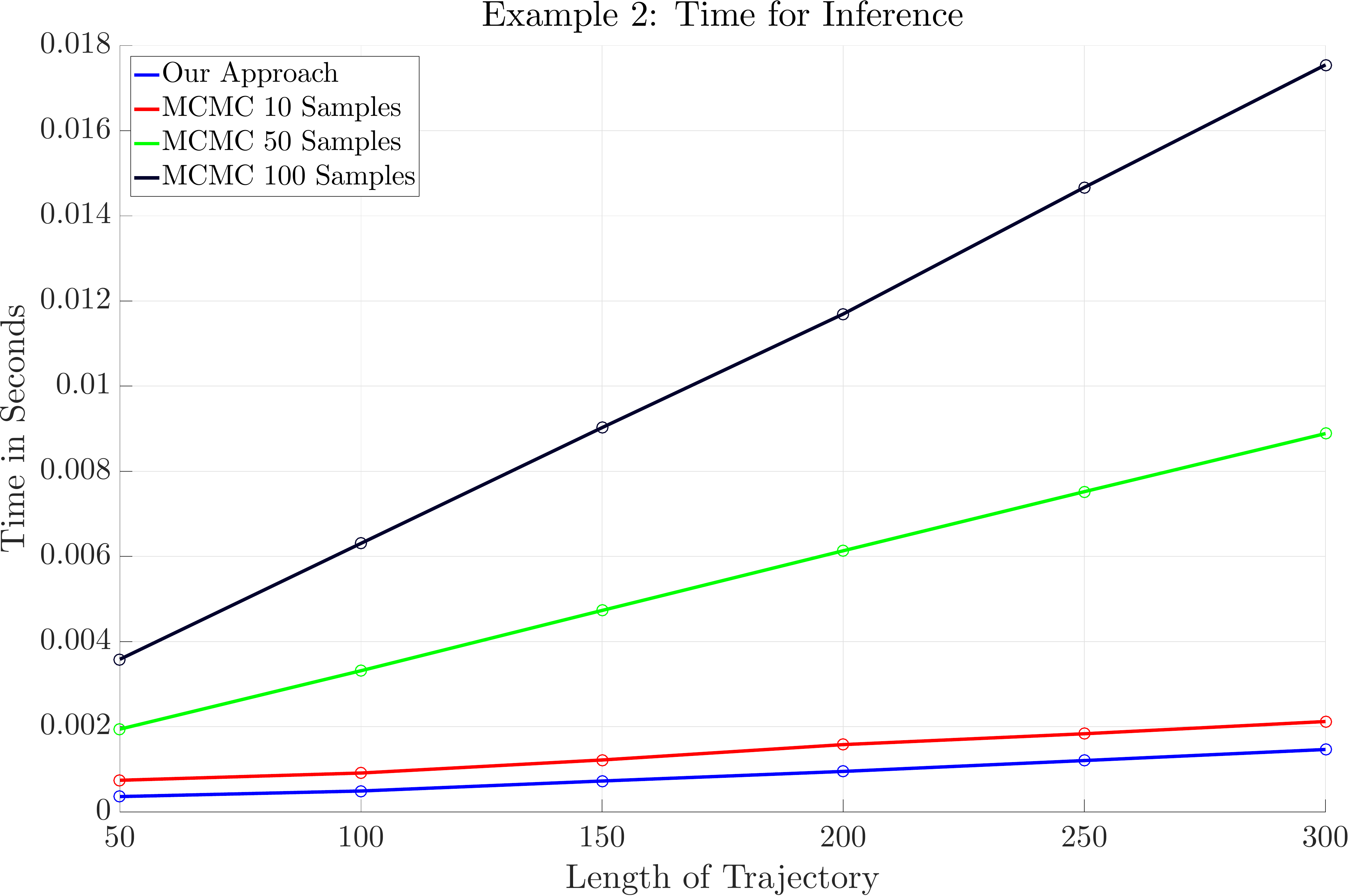}
	\label{blah2}
\end{subfigure}
~
\begin{subfigure}[b]{0.3\textwidth}
	\includegraphics[width=\textwidth]{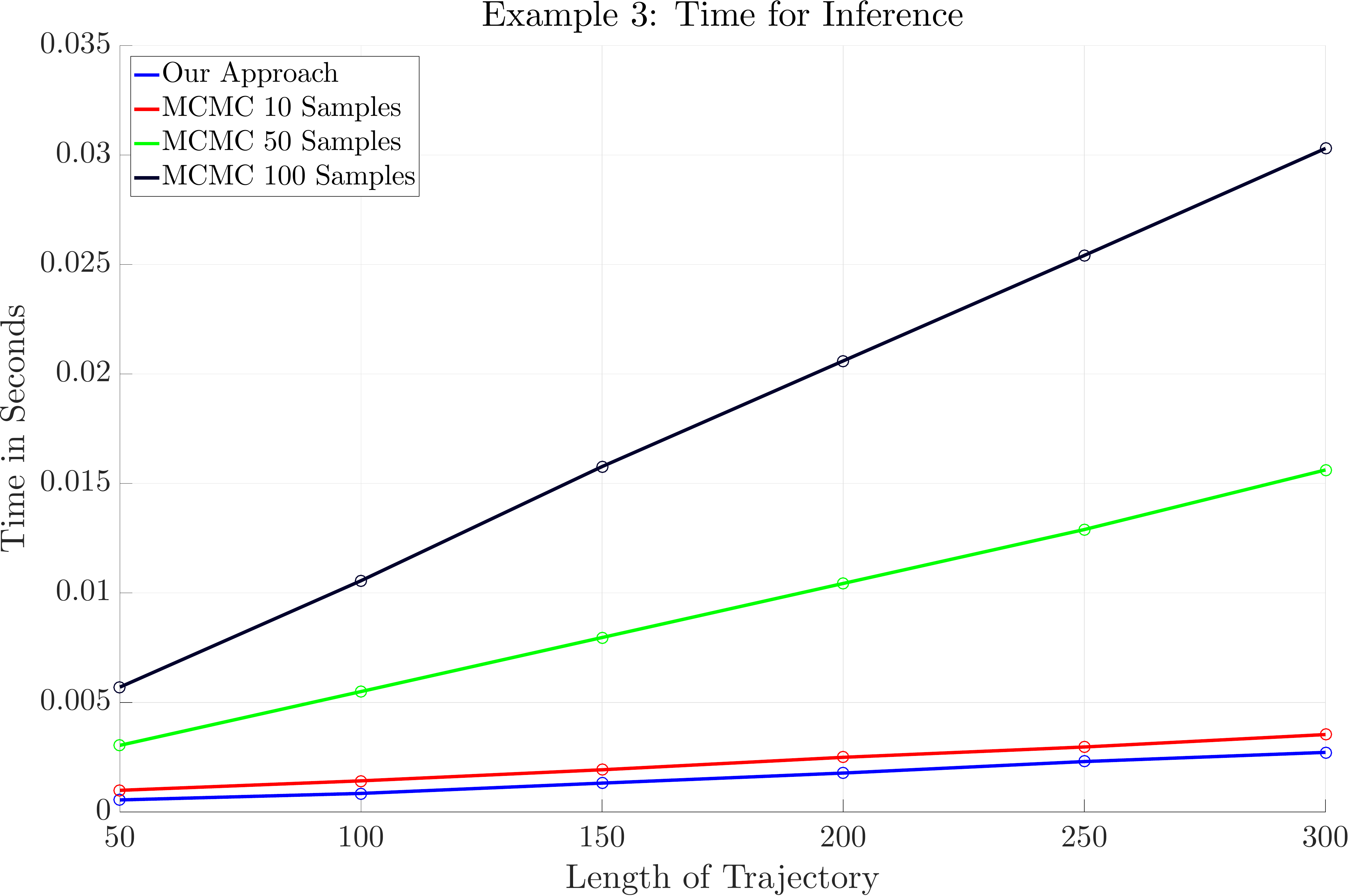}
	\label{blah3}
\end{subfigure}
\begin{subfigure}[b]{0.3\textwidth}
	\includegraphics[width=\textwidth]{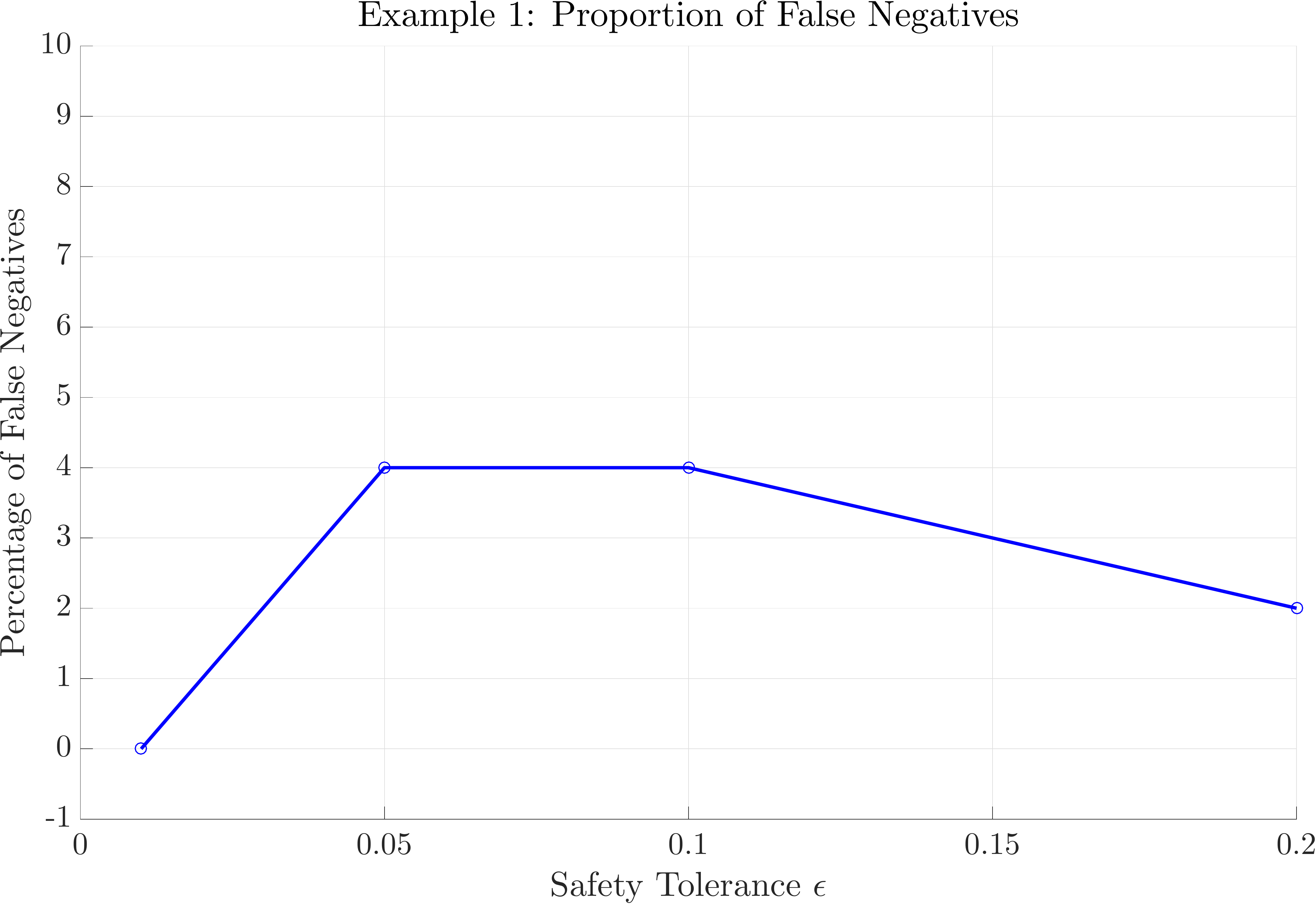}
	\label{blaht1}
\end{subfigure}
~
\begin{subfigure}[b]{0.3\textwidth}
	\includegraphics[width=\textwidth]{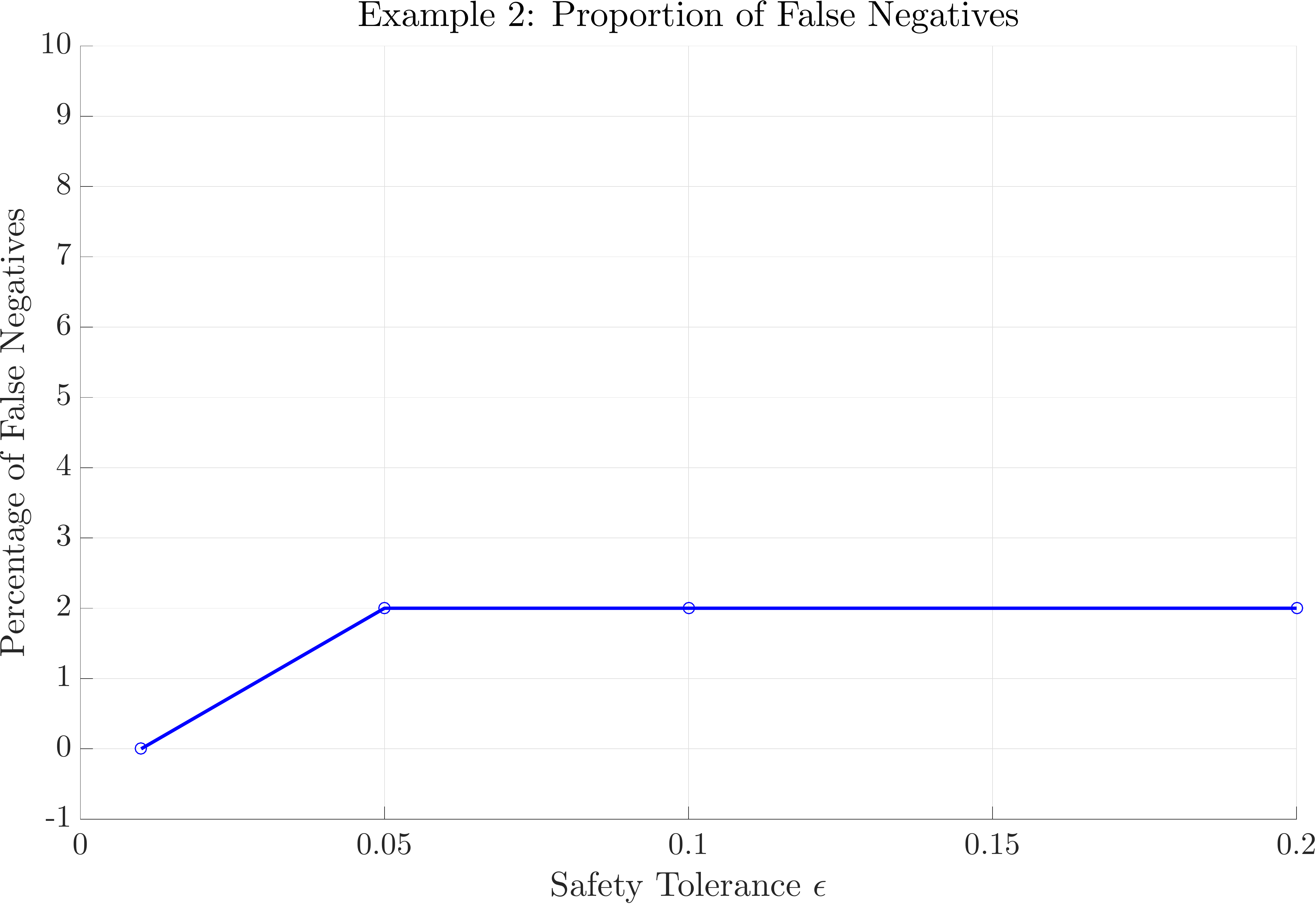}
	\label{blaht2}
\end{subfigure}
~
\begin{subfigure}[b]{0.3\textwidth}
	\includegraphics[width=\textwidth]{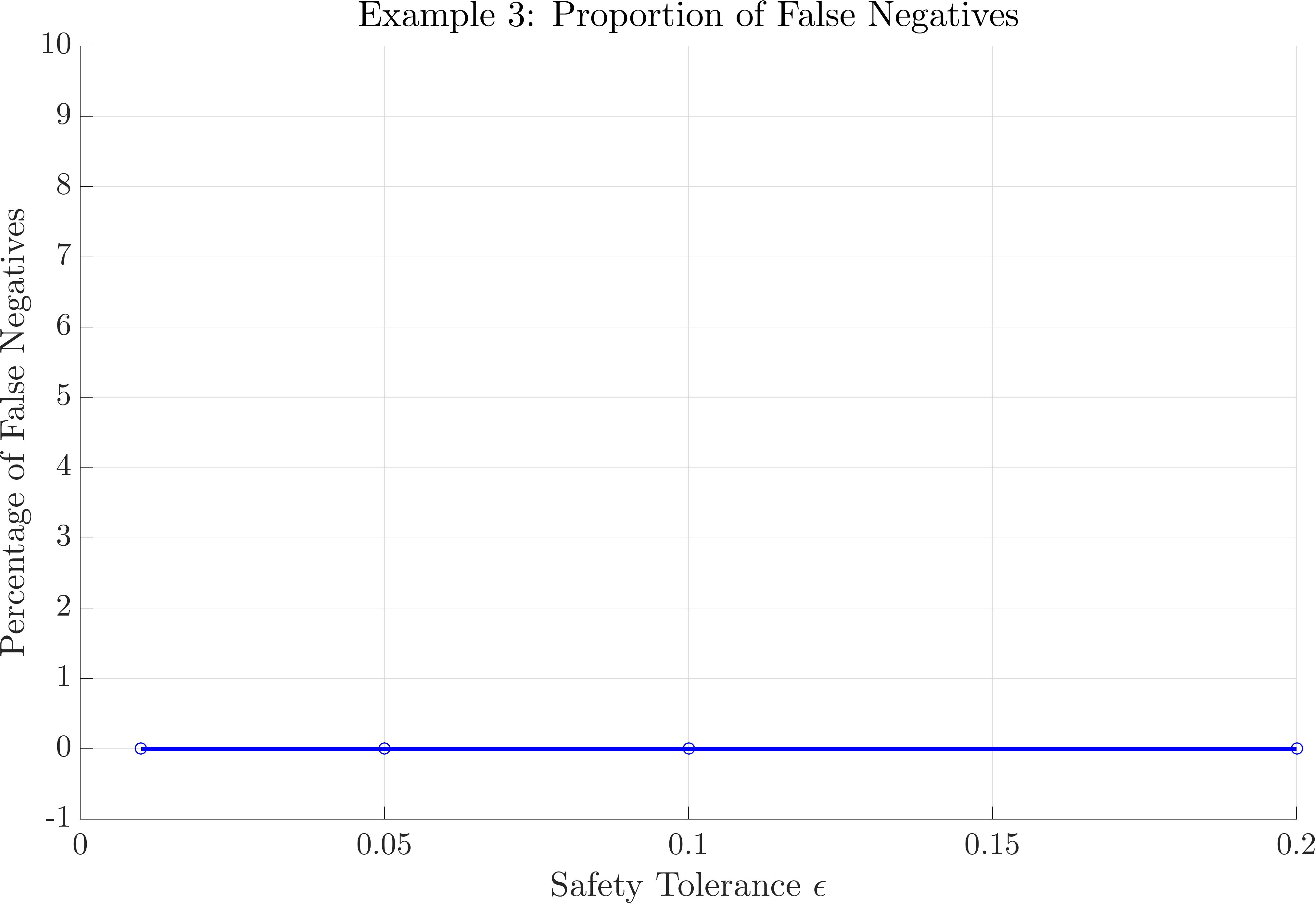}
	\label{blaht3}
\end{subfigure}
\caption{Performance comparison of the proposed inference algorithm. The top row depicts the completion time of running the three procedures as the length of the trajectory is varied. The bottom row shows how many safe trajectories does the approximate inference procedure misclassifies as unsafe. We observe that the proposed strategy is fast as well as accurate.}
\end{figure*}

\section{Examples and Experiments}
We perform experiments on three different probabilistic safety programs motivated by real-world applications. The goal
of these experiments is to explore both the efficiency as well as accuracy of estimating the safety query. In addition, we
also provide a real-world example of how a PSP embedded in an RRT* planner can be effectively used to build quadrotor systems that are capable of safe flight
in the presence of unknown obstacles.

\begin{figure*}
\centering
\includegraphics[width=\textwidth]{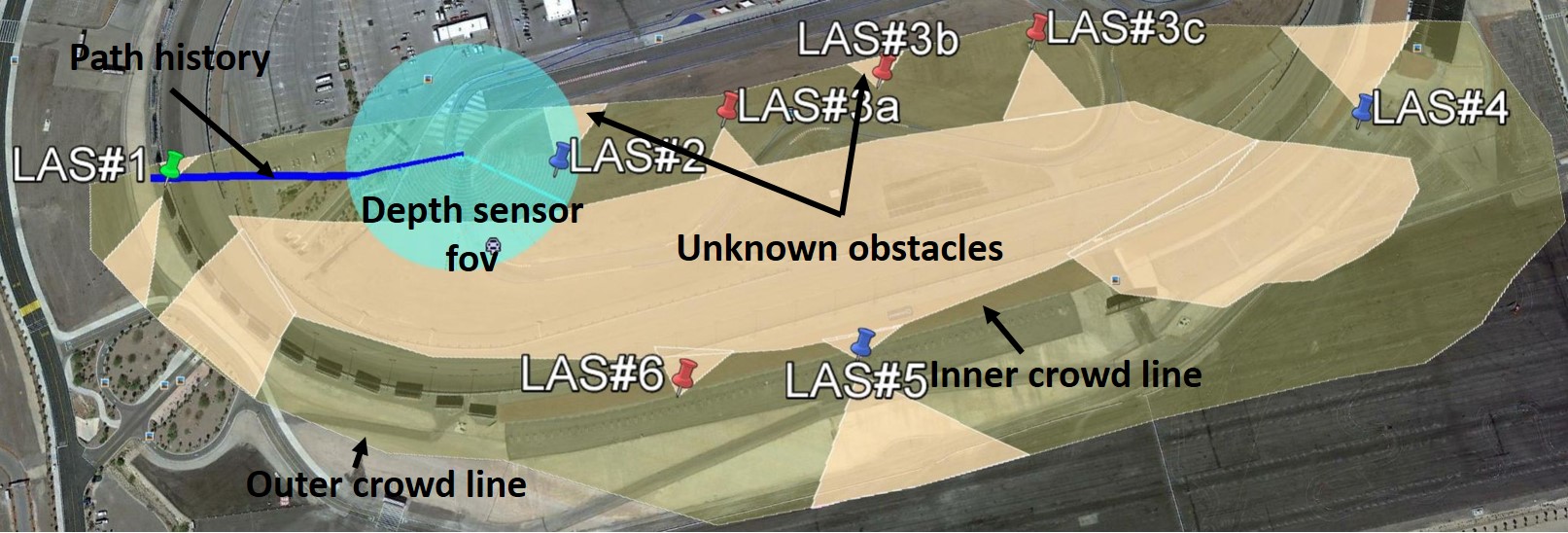}
\caption{Las Vegas Motor Speedway Air Race course used to showcase safe planning for quadrotors.}
\label{course_annotated}
\end{figure*}

Our three simulation examples include the following scenarios:

\noindent{\bf Example 1:} This is the same scenario as described earlier in Fig. 1, where the goal is to avoid obstacles via a machine learning system characterized as a Bayesian linear classifier.

\noindent{\bf Example 2:} The goal in this example is to address quadrotor flight safety then there is uncertainty about the battery level. The battery level is a stochastic
variable due to uncertain environment and usage factors, such as radio communications, etc. Our goal is to derive a safe strategy that considers such stochasticity. Specifically, we want to avoid flights
above an altitude \texttt{heightThresh} of we do not have enough battery (less than \texttt{batteryThresh}). Thus, the PSP shown in Figure 2 first samples from a predictive distribution of the battery level and enforces the safety constraint that only those trajectories are allowed that has the minimum battery level at least three steps before the quadrotor assumes a height greater than the threshold. Note that the PSP assumes that the uncertainty in the battery level grows linearly with elapsed time steps. This is due to the fact that the variance of the Gaussian distribution from which \texttt{batteryNow} is being sampled is expressed as \texttt{i*variance}.

\noindent{\bf Example 3:} The third example illustrates the use of PSPs in autonomous driving. Assuming that there is another car on the road with whom we need to maintain safe distance, we write a PSP shown in Figure 3 right that returns true only if for all the positions of the ego vehicle, the distance to the other car is greater than \texttt{Threshold} either in x or y direction. In this example, we assume that we have a predictive model corresponding to the other car that provides probability distribution over both its future positions and speeds. In line with the previous example, this PSP first samples positions and velocities from the predictive distribution and then evaluates the safety invariant.

For our experiments we statically compile each of these probabilistic programs to a Bayesian graphical model as mentioned in section 3. Next, we generate $50$ random parameters of the probability distribution that are
responsible for the stochasticity in the programs (e.g. ${\mathbf \mu}$ and ${\mathbf \Sigma}$ in example 1 etc.). For each of the randomly generated parameters we run the approximate inference algorithm on a set of pre-computed trajectories. Besides running the approximate inference, we also run inference in this models via a sampling procedure based on Markov Chain Monte Carlo (MCMC).

Figure 4 top row shows the running time of the two inference algorithms for all the three different examples. The x-axis is the size of the trajectory for which the safety query is invoked, whereas the y-axis indicates the actual running time. Besides plotting the running time for the approximate inference, we also plot the curves for MCMC procedure with different number of samples. We observe that the approximate inference procedure has significant advantages in terms of the running time for all the three PSPs. Furthermore, this advantage gets even more significant as the length of the trajectory is increased. Note that we were able to run the approximate inference engine in less than 0.003 secs on average for trajectories of length 300, highlighting that framework can be run over 300 Hz.

Besides the running time we are also interested in the accuracy of the approximate inference model. As we have seen in Section 3 the approximate inference method bounds Pr(\texttt{Program is Safe}) from below. So while it is impossible for the approach to infer an unsafe trajectory as safe, it can still infer a safe trajectory as unsafe. Figure 4 bottom row highlights the percentage of trajectories that are false negatives (i.e. safe trajectories characterized as unsafe) as we vary the threshold parameter $\epsilon$. We also plot these curves for trajectories of different lengths. The figures highlight that if use the approximate inference then the proportion of false positives are less than 4\% for all the cases that we ran. These results indicate that the approximate inference procedure is both computationally efficient and highly accurate.

\noindent{\bf Online Safe Planning for Quadrotors with PSPs:} Next, we showcase how PSPs can be effectively embedded in existing planners. We tackle an online planning scenario for a quadrotor which is tasked with flying through an obstacle course which can have known as well as \emph{unknown} obstacles. Figure \ref{course_annotated} shows the Las Vegas Motor Speedway Air Race course used for this task. We assume that the quadrotor is equipped with a spinning lidar sensor that is capable of returning depth to nearest obstacles in its immediate vicinity (blue circle). Consequently, our probabilistic safety program looks like Example 1 provided earlier, where the observations from lidar are used to build a Bayesian predictive model and then used in the PSP for safety checks.

The quadrotor is tasked with starting at the first gate (``LAS\#1'') and must reach gate ``LAS\#6'' while traversing through all the gates in sequence. In addition to avoiding known and unknown obstacles it must also obey rules specific to the course. These include maintaining minimum altitude of $75$ feet and staying inside the outer crowd line and outside the inner crowd line. We take a receding-horizon \cite{howard2010receding} approach
 with RRT* \cite{karaman2011sampling}, a sampling based motion planner that works by constructing a tree via random samples of the workspace. Sampled nodes are connected to the tree, only if collision-free trajectories can be found to them from the nearest node in the tree. These collision checks for the unknown obstacles are performed via the Probabilistic Safety Program framework.

\begin{figure*}	
	\centering

	\begin{subfigure}[t]{0.5\textwidth}
		\centering
		\includegraphics[width=\textwidth]{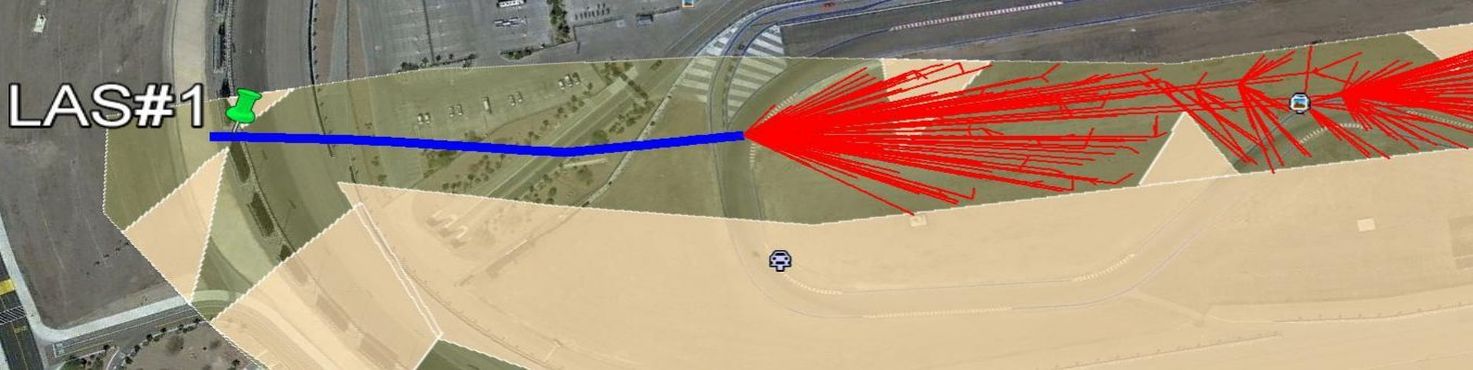}
		\caption{RRT* tree (red) generated from current position.}
        \label{course_unsafe}
	\end{subfigure}%
    \begin{subfigure}[t]{0.5\textwidth}
		\centering
		\includegraphics[width=\textwidth]{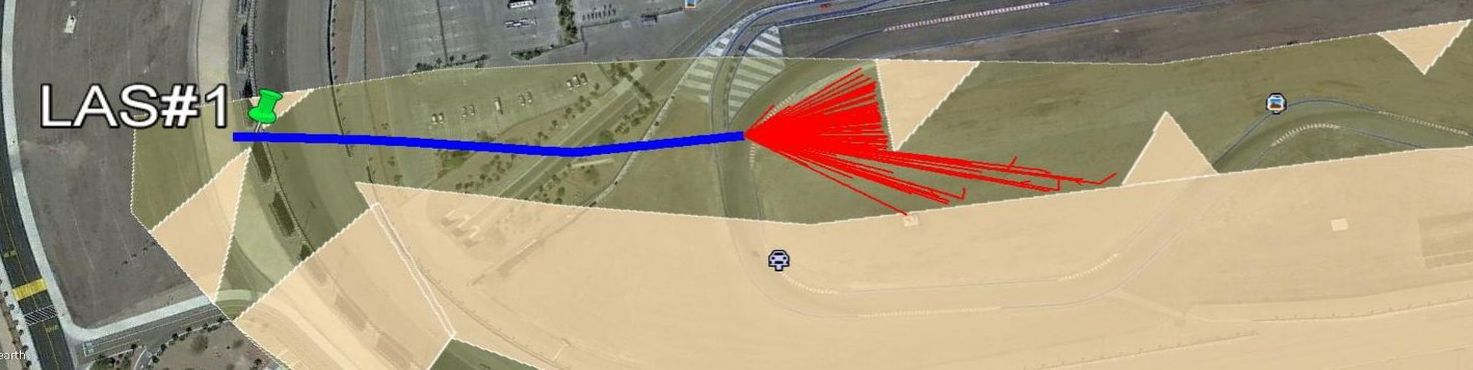}
		\caption{Using PSP to prune unsafe regions of the tree.}
        \label{course_safe}
	\end{subfigure}
    \begin{subfigure}[t]{0.5\textwidth}
		\centering
		\includegraphics[width=\textwidth]{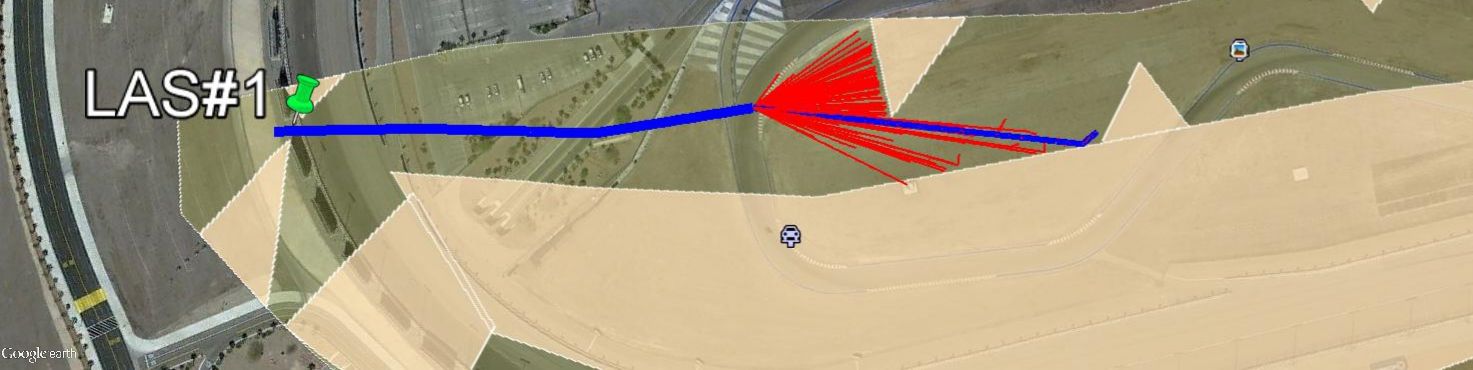}
		\caption{Shortest path in safe tree found towards goal node.}
        \label{course_safe_short}
	\end{subfigure}%
    \begin{subfigure}[t]{0.5\textwidth}
		\centering
		\includegraphics[width=\textwidth]{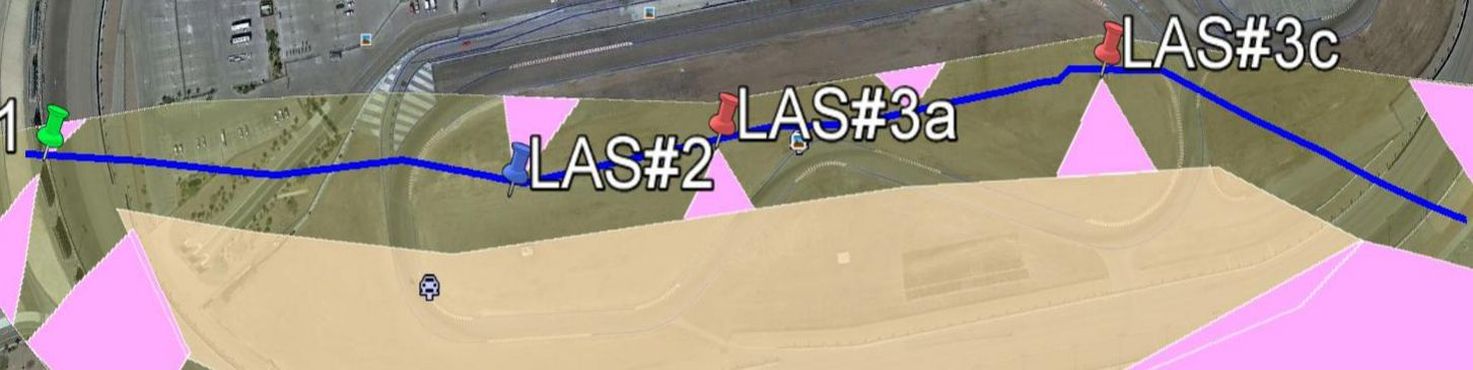}
		\caption{Traversed trajectory from first gate to the fourth gate.}
        \label{course_complete}
	\end{subfigure}

	\caption{The PSP when embedded in a planner based on RRT* helps avoid unknown obstacles.}\label{fig:1}
\end{figure*}

Figure \ref{course_unsafe} shows the generated RRT* tree from the current location of the quadrotor. While the tree takes into account known obstacles and other course-specific safety rules, without PSP it passes through the unknown obstacle right in front of it. Figure \ref{course_safe} shows the RRT* tree pruned using the PSP, which avoids the previously unknown obstacle thanks to the PSP check. Figure \ref{course_safe_short} the shortest path to the goal location is determined by Dijkstra's algorithm \cite{dijkstra1959note} (indicated by the blue line) and then partially traversed by the quadrotor. This process is then repeated from the new location until the goal is reached. The depth sensor has a radius of $90$ meters. $300$ valid nodes were sampled by RRT* for each cycle detailed above. Each edge in the RRT* tree had to satisfy at least greater than $50$\% chance of being collision-free to be kept. Figure \ref{course_complete} shows the complete path taken by the quadrotor till the fourth gate while successfully avoiding the two unknown obstacles along the way.

\section{Related Work}
Related to this work are several temporal logic specification languages that have been developed and adapted for synthesizing controllers such as Linear Temporal Logic (LTL) \cite{kress2009temporal,tabuada2004linear}, Metric Temporal Logic (MTL) \cite{karaman2008optimal}, Probabilistic Temporal Logic (PTL) \cite{yoo2012probabilistic} and Signal Temporal Logic (STL) \cite{raman2015reactive,raman2014model,maler2004monitoring}. These approaches can be used for task planning \cite{plakumotion}, where a system designer a priori specifies logical specifications composed of disjunctions, conjunctions, negations as well as temporal permutations of those combinations. Previous work has also proposed methods for combining sampling-based motion planners with such specification languages to do joint task and motion planning for autonomous robots \cite{karaman2009sampling,karaman2008optimal}. However, these approaches are limited in their capacity to both express the constraints as well as the capability to account for uncertainty in sensors and dynamics.

Also there have been previous efforts to incorporate such uncertainty into planning \cite{melchior2007particle,deyVisionQuad2015,van2011motion}, however, safety guarantees have been difficult to provide in such settings especially under constrained computational budget. One of the closest approaches to our work is that of \cite{luders2010chance} who propose a chance-constrained RRT (CC-RRT) where uncertainty in dynamic obstacles and sensing is propagated down the tree and only those paths in the tree are kept such that they satisfy a real-time constraint. In contrast to chance-constraints, Probabilistic Safety Programs allows a much richer class of Boolean and temporal constraints to be specified under uncertainty.

\section{Conclusion and Future Work}
We have presented Probabilistic Safe Programs that enables specification and inference over safety constraints when the system is operating in an uncertain environment. We ensure their satisfaction via efficient reductions to equivalent graphical models and performing inference via fast approximate but bounded Bayesian inference. We have demonstrated the efficiency of the proposed techniques via multiple examples and in conjunction with widely used planners in real-world systems. In future, we aim to deploy these on real robots with multiple sensing modalities and further integrate them with deliberative planners to help close the gap between perception and planning.

\bibliographystyle{IEEEtran}
\bibliography{IEEEabrv,references}

\end{document}

%% file: macros.tex
\usepackage{graphicx,caption,subcaption}
\usepackage{amsmath,amssymb,stmaryrd,mathtools}
\usepackage{flushend}
\usepackage{hyperref}
\usepackage{acronym}

\graphicspath{{./figures/}}

\newcommand{\ignore}[1]{}

\DeclareFontFamily{U}{MnSymbolC}{}
\DeclareSymbolFont{MnSyC}{U}{MnSymbolC}{m}{n}
\DeclareFontShape{U}{MnSymbolC}{m}{n}{
    <-6>  MnSymbolC5
   <6-7>  MnSymbolC6
   <7-8>  MnSymbolC7
   <8-9>  MnSymbolC8
   <9-10> MnSymbolC9
  <10-12> MnSymbolC10
  <12->   MnSymbolC12%
}{}
\DeclareMathSymbol{\powerset}{\mathord}{MnSyC}{180}


%% file: root.bbl
\begin{thebibliography}{10}
\providecommand{\url}[1]{#1}
\csname url@rmstyle\endcsname
\providecommand{\newblock}{\relax}
\providecommand{\bibinfo}[2]{#2}
\providecommand\BIBentrySTDinterwordspacing{\spaceskip=0pt\relax}
\providecommand\BIBentryALTinterwordstretchfactor{4}
\providecommand\BIBentryALTinterwordspacing{\spaceskip=\fontdimen2\font plus
\BIBentryALTinterwordstretchfactor\fontdimen3\font minus
  \fontdimen4\font\relax}
\providecommand\BIBforeignlanguage[2]{{%
\expandafter\ifx\csname l@#1\endcsname\relax
\typeout{** WARNING: IEEEtran.bst: No hyphenation pattern has been}%
\typeout{** loaded for the language `#1'. Using the pattern for}%
\typeout{** the default language instead.}%
\else
\language=\csname l@#1\endcsname
\fi
#2}}

\bibitem{urmson2008autonomous}
C.~Urmson, J.~Anhalt, D.~Bagnell, C.~Baker, R.~Bittner, M.~Clark, J.~Dolan,
  D.~Duggins, T.~Galatali, C.~Geyer, \emph{et~al.}, ``Autonomous driving in
  urban environments: Boss and the urban challenge,'' \emph{Journal of Field
  Robotics}, 2008.

\bibitem{GMR+08}
N.~D. Goodman, V.~K. Mansinghka, D.~M. Roy, K.~Bonawitz, and J.~B. Tenenbaum,
  ``Church: a language for generative models,'' in \emph{UAI}, 2008.

\bibitem{GHNR14}
A.~D. Gordon, T.~A. Henzinger, A.~V. Nori, and S.~K. Rajamani, ``Probabilistic
  programming,'' in \emph{ICSE, FOSE track}, 2014.

\bibitem{MSS09}
A.~McCallum, K.~Schultz, and S.~Singh, ``Factorie: Probabilistic programming
  via imperatively defined factor graphs,'' in \emph{NIPS}, 2009.

\bibitem{kulkarni2015picture}
T.~D. Kulkarni, P.~Kohli, J.~B. Tenenbaum, and V.~Mansinghka, ``Picture: A
  probabilistic programming language for scene perception,'' in \emph{CVPR},
  2015.

\bibitem{BLOG}
\BIBentryALTinterwordspacing
B.~Milch, B.~Marthi, S.~Russell, D.~Sontag, D.~L. Ong, and A.~Kolobov,
  ``{BLOG}: Probabilistic models with unknown objects,'' in \emph{Statistical
  Relational Learning}, L.~Getoor and B.~Taskar, Eds.\hskip 1em plus 0.5em
  minus 0.4em\relax MIT Press, 2007. [Online]. Available:
  \url{http://sites.google.com/site/bmilch/papers/blog-chapter.pdf}
\BIBentrySTDinterwordspacing

\bibitem{howard2010receding}
T.~M. Howard, C.~J. Green, and A.~Kelly, ``Receding horizon model-predictive
  control for mobile robot navigation of intricate paths,'' in \emph{Field and
  Service Robotics}, 2010.

\bibitem{lavalle2000rapidly}
S.~M. Lavalle and J.~J. Kuffner~Jr, ``Rapidly-exploring random trees: Progress
  and prospects,'' in \emph{Algorithmic and Computational Robotics: New
  Directions}, 2000.

\bibitem{karaman2011sampling}
S.~Karaman and E.~Frazzoli, ``Sampling-based algorithms for optimal motion
  planning,'' \emph{The International Journal of Robotics Research}, 2011.

\bibitem{WGSS11}
D.~Wingate, N.~D. Goodman, A.~Stuhlm\"uller, and J.~M. Siskind, ``Nonstandard
  interpretations of probabilistic programs for efficient inference,'' in
  \emph{NIPS}, 2011.

\bibitem{SWSG15}
A.~Stuhlm\"{u}ller, R.~X.~D. Hawkins, N.~Siddharth, and N.~D. Goodman,
  ``Coarse-to-fine sequential monte carlo for probabilistic programs,'' 2015.

\bibitem{WVM14}
F.~Wood, J.~W. van~de Meent, and V.~Mansinghka, ``A new approach to
  probabilistic programming inference,'' in \emph{Artificial Intelligence and
  Statistics}, 2014.

\bibitem{sadighPrSTL2015}
D.~Sadigh and A.~Kapoor, ``Safe control under uncertainty,'' in \emph{Robotics:
  Science and Systems}, 2016.

\bibitem{dijkstra1959note}
E.~W. Dijkstra, ``A note on two problems in connexion with graphs,''
  \emph{Numerische mathematik}, vol.~1, no.~1, pp. 269--271, 1959.

\bibitem{kress2009temporal}
H.~Kress-Gazit, G.~E. Fainekos, and G.~J. Pappas, ``Temporal-logic-based
  reactive mission and motion planning,'' \emph{Robotics, IEEE Transactions
  on}, 2009.

\bibitem{tabuada2004linear}
P.~Tabuada and G.~J. Pappas, ``Linear temporal logic control of linear
  systems,'' \emph{IEEE Transactions on Automatic Control}, 2004.

\bibitem{karaman2008optimal}
S.~Karaman and E.~Frazzoli, ``Optimal vehicle routing with metric temporal
  logic specifications,'' in \emph{CDC}, 2008.

\bibitem{yoo2012probabilistic}
C.~Yoo, R.~Fitch, and S.~Sukkarieh, ``Probabilistic temporal logic for motion
  planning with resource threshold constraints,'' in \emph{RSS}, 2012.

\bibitem{raman2015reactive}
V.~Raman, A.~Donz{\'e}, D.~Sadigh, R.~M. Murray, and S.~A. Seshia, ``Reactive
  synthesis from signal temporal logic specifications,'' in \emph{Proceedings
  of the 18th International Conference on Hybrid Systems: Computation and
  Control}.\hskip 1em plus 0.5em minus 0.4em\relax ACM, 2015.

\bibitem{raman2014model}
V.~Raman, A.~Donz{\'e}, M.~Maasoumy, R.~M. Murray, A.~Sangiovanni-Vincentelli,
  and S.~A. Seshia, ``Model predictive control with signal temporal logic
  specifications,'' in \emph{CDC}, 2014.

\bibitem{maler2004monitoring}
O.~Maler and D.~Nickovic, ``Monitoring temporal properties of continuous
  signals,'' in \emph{Formal Techniques, Modelling and Analysis of Timed and
  Fault-Tolerant Systems}, 2004.

\bibitem{plakumotion}
E.~Plaku and S.~Karaman, ``Motion planning with temporal-logic specifications:
  Progress and challenges,'' \emph{AI Communications}, 2015.

\bibitem{karaman2009sampling}
S.~Karaman and E.~Frazzoli, ``Sampling-based motion planning with deterministic
  $\mu$-calculus specifications,'' in \emph{Decision and Control}, 2009.

\bibitem{melchior2007particle}
N.~A. Melchior and R.~Simmons, ``Particle rrt for path planning with
  uncertainty,'' in \emph{ICRA}, 2007.

\bibitem{deyVisionQuad2015}
D.~Dey, K.~S. Shankar, S.~Zeng, R.~Mehta, M.~T. Agcayazi, C.~Eriksen,
  S.~Daftry, M.~Hebert, and J.~A. Bagnell, ``Vision and learning for
  deliberative monocular cluttered flight,'' \emph{Field and Service Robotics},
  2015.

\bibitem{van2011motion}
J.~Van Den~Berg, S.~Patil, and R.~Alterovitz, ``Motion planning under
  uncertainty using differential dynamic programming in belief space,'' in
  \emph{Int’l Symposium on Robotics Research}, 2011.

\bibitem{luders2010chance}
B.~Luders, M.~Kothari, and J.~P. How, ``Chance constrained rrt for
  probabilistic robustness to environmental uncertainty,'' in \emph{AIAA
  guidance, navigation, and control conference (GNC)}, 2010.

\end{thebibliography}
